\documentclass[a4paper,10pt]{article}

\usepackage[T1]{fontenc}
\usepackage[utf8]{inputenc}

\usepackage{graphicx}
\usepackage{tabu}
\usepackage{multirow}
\usepackage{multicol}
\usepackage{ragged2e}
\usepackage{url}
\usepackage{latexsym}
\usepackage{booktabs}
\usepackage{amsmath,amsfonts}
\usepackage{braket}
\usepackage[breaklinks=true]{hyperref}
\usepackage{xcolor}
\usepackage{colortbl}
\usepackage{lipsum}
\usepackage{enumitem}
\usepackage{microtype}

\usepackage{caption}

\usepackage{iwslt18}

\newcommand{\tbf}[1]{\textbf{#1}}
\newcommand{\ttt}[1]{\texttt{#1}}

\newcommand{\tsub}[1]{\textsubscript{#1}}

\newcommand{\citep}[1]{\cite{#1}}
\newcommand{\citet}[1]{\cite{#1}}

\title{The LIG system for the English-Czech Text Translation Task of IWSLT 2019}
\name{\parbox{\linewidth}{\centering Loïc Vial \quad Benjamin Lecouteux \quad Didier Schwab\smallskip\\ Hang Le \quad Laurent Besacier}}
\address{\parbox{\linewidth}{\centering Univ. Grenoble Alpes, CNRS, Grenoble INP, LIG, 38000 Grenoble, France\smallskip\\
\texttt{\{loic.vial, benjamin.lecouteux, didier.schwab,\\ thi-phuong-hang.le, laurent.besacier\}@univ-grenoble-alpes.fr}}}

\begin{document}

\maketitle

\begin{abstract}
    In this paper, we present our submission for the English to Czech Text Translation Task of IWSLT 2019. Our system aims to study how pre-trained language models, used as input embeddings, can improve a specialized machine translation system trained on few data.

    Therefore, we implemented a Transformer-based encoder-decoder neural system which is able to use the output of a pre-trained language model as input embeddings, and we compared its performance under three configurations: 1) without any pre-trained language model (constrained), 2) using a language model trained on the monolingual parts of the allowed English-Czech data (constrained), and 3) using a language model trained on a large quantity of external monolingual data (unconstrained). 
    We used BERT as external pre-trained language model (configuration 3), and BERT architecture for training our own language model (configuration 2).
    
    Regarding the training data, we trained our MT system on a small quantity of parallel text: one set only consists of the provided MuST-C corpus, and the other set consists of the MuST-C corpus and the News Commentary corpus from WMT. 
    
    We observed that using the external pre-trained BERT improves the scores of our system by +0.8 to +1.5 of BLEU on our development set, and +0.97 to +1.94 of BLEU on the test set. However, using our own language model trained only on the allowed parallel data seems to improve the machine translation performances only when the system is trained on the smallest dataset. 
\end{abstract}

\section{Introduction}

The recent advances in pre-trained Language Models \cite{Peters2018,devlin2018bert,lample2019crosslingual,yang2019xlnet,liu2019roberta} have shown that they could greatly improve many NLP tasks such as Natural Language Understanding, Question Answering, Natural Language Inference, Word Sense Disambiguation, etc.

With our submission, we would like to explore what these models can bring to a typical Transformer-based encoder-decoder Neural Machine Translation system.
Unlike the works of \citep{ramachandran-etal-2017-unsupervised} and \citep{lample2019crosslingual} where the authors fine-tune the weights of the language models on the translation task, we propose to use the language models as input embeddings for our neural system. 

We expect that the language model, because it is trained on a great quantity of monolingual data, will bring some additional information to a MT system trained on relatively few parallel data.

Therefore, we conducted experiments that compare our system with and without the information from a BERT pre-trained model \citep{devlin2018bert}. In addition, we created our own BERT LM by training it on the allowed training data only, in order to see if the language model is still useful in a constrained setting.

For the training data, we used only the provided \mbox{MuST-C} \citep{mustc19} and News Commentary \citep{barraultEtAl2019WMT} from WMT, for a total of less than 400k parallel sentences.

\section{System Description}

\subsection{Architecture}

\begin{figure}[htbp]
    \centering
    \includegraphics[width=1.0\linewidth]{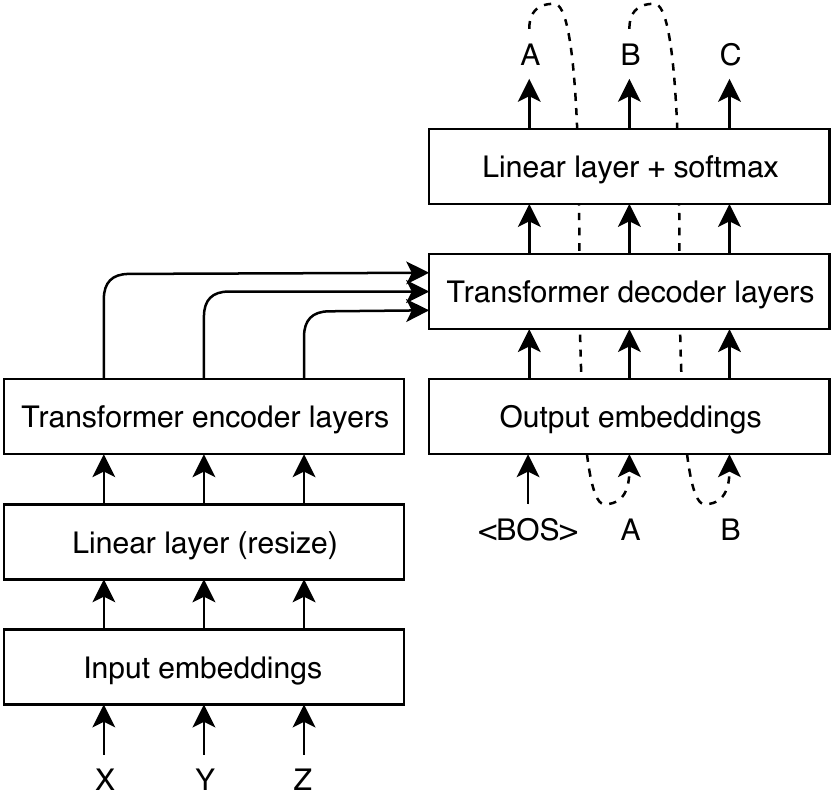}
    \caption{Architecture of our neural MT system.}
    \label{fig:network}
\end{figure}

Our system relies mostly on the Transformer architecture \citep{vaswani2017}. More precisely, it consists of the following layers, as pictured in \autoref{fig:network}:
\begin{itemize}
    \item The input embeddings layer, which takes words in their vector form from either 1) a classical look-up table trained jointly with the model or 2) a pre-trained language model which remains fixed during the training. 
    \item A linear layer, only if the embeddings come from a pre-trained language model, in order to resize their vectors to the desired size.
    \item Multiple Transformer encoder layers. 
    \item The output embeddings layer (trained look-up table). 
    \item Multiple Transformer decoder layers.
    \item A linear layer which resizes the decoder output to the output vocabulary size, followed by a softmax.
\end{itemize}

We implemented our system using PyTorch\footnote{\url{https://pytorch.org/}}.
For the Transformer encoder and decoder layers, we used the implementation from OpenNMT\footnote{\url{https://github.com/OpenNMT/OpenNMT-py}}. The parameters used are the same as the ``base'' model of \cite{vaswani2017}: 6 layers, 8 attention heads and a hidden feed-forward size of 2048, except for the dropout rate that we set to $0.3$ to improve the robustness of our model. 

It is to be noted that, as in \citep{vaswani2017}, we share the weight matrix between the output embeddings and the last linear layer. However, we do not share the vocabulary nor the matrices between the input and the output languages.

Also, for the input embeddings, if they come from a look-up table, we add sinusoidal positional encoding to the vectors as in \citep{vaswani2017}. We do not need it when using a language model because the positions are already encoded.

Finally, for the size of the embeddings, which is the same as the input and output of the Transformer layers, we tried two different parameters: 512 and 1024.

\subsection{Training and development corpora}

Due to time constraints, we limited our training to only two English-Czech corpora that we considered of good quality and relevant for the task: the provided MuST-C \citep{mustc19} and the News Commentary corpus provided by WMT \citep{barraultEtAl2019WMT}. MuST-C is a speech translation corpus of TED talks, similar to the test data of the task, and we added the News Commentary corpus, which consists of political and economic commentaries, because it was the second smallest corpus provided by WMT, after Common Crawl, and we estimated that its quality was better than Common Crawl. 


The training data of MuST-C contains 128\,179 sentences, and the News Commentary corpus contains 246\,513 sentences. We conducted two sets of experiments: one using only the MuST-C, and the other using both corpora, hence cumulating 374\,692 training sentences.


For the development set used in both settings, we used the development and test corpora from MuST-C, which corresponds to 3\,928 sentences.

\subsection{Preprocessing}

We preprocessed every corpus using the standard scripts from the Moses repository.\footnote{\url{https://github.com/moses-smt/mosesdecoder}} In particular, we normalized punctuation characters,
removed non-printing characters,
and tokenized the data.
Finally, we removed sentences with more than 80 words and those with a source-target word ratio greater than $1.5$.



\begin{table}[htbp]
\centering
\tabulinesep=3pt
\taburowcolors[2]3{white..gray!10}
\begin{tabu} to \linewidth {|X[2lm]|X[2cm]|X[2cm]|}
\firsthline
Corpus & Original sentence count & Cleaned sentence count \\
\hline
MuST-C & 128\,179 & 112\,993 \\
MuST-C + News & 374\,692 & 344\,973\\
Dev
& 3\,928 & 3\,507\\
\lasthline
\end{tabu}
\caption{Corpora statistics before and after cleaning.}
\label{tab:training_corpora}
\end{table}

\autoref{tab:training_corpora} summarize the corpus lengths before and after the preprocessing phase.

\subsection{Vocabulary}

\subsubsection{English side}

There are tree cases for the input English vocabulary:
\begin{enumerate}[topsep=3pt,itemsep=3pt,parsep=1pt,partopsep=1pt]
    \item For the case where we used BERT external pre-trained language model, we use the same vocabulary as their model named ``bert-base-cased'' which consists of 30\,000 subwords.
    \item For the case where we trained our own BERT constrained model, we used a BPE vocabulary of size 30\,000 trained on all allowed corpora for the task, which consists of MuST-C and 6 other corpora from WMT\footnote{\url{http://www.statmt.org/wmt19/translation-task.html}}.
    \item For the case where we do not use any language model, we trained a BPE vocabulary of size 30\,000, but only on MuST-C and News Commentary.
\end{enumerate}

\subsubsection{Czech side}

For the output Czech vocabulary, we used the same in every configuration: we learned a BPE vocabulary of size 14\,000 on the Czech side of the MuST-C and the News Commentary corpora.
For BPE learning, we used the tool subword-nmt\footnote{\url{https://github.com/rsennrich/subword-nmt}}.

\subsection{Language model pre-training}

In order to both be able to explore how much a pre-trained language model can improve a NMT system, and submit a system constrained in terms of training data, we trained our own language model restricted to the allowed data. 


We used the English side of the corpora listed in \autoref{tab:training_language_model} for the pre-training data, and we extracted $0.5\%$ of the sentences for the validation and test sets (approximately 314\,000 sentences). 
Comparing to the BERT external pre-trained model ``bert-base-cased'' provided by the authors, which is trained on a corpus set that contains more than 3 billions words, we have 708\,622\,867 words in total which amounts to approximately 20\%.

\begin{table}[htbp]
\centering
\tabulinesep=3pt
\taburowcolors[2]3{white..gray!10}
\begin{tabu} to \linewidth {|X[2.7lm]|X[2cm]|X[2cm]|}
\firsthline
Corpus & Sentence count & Word count \\
\hline
MuST-C & 128\,179 & 2\,413\,793 \\
News Commentary & 246\,513 & 5\,168\,469 \\
Common Crawl & 161\,838 & 3\,348\,584\\
Wiki Titles & 362\,015 & 897\,564\\
Europarl & 654\,323 & 15\,628\,367 \\
ParaCrawl & 2\,981\,949 & 48\,918\,150 \\
CzEng & 58\,315\,645 & 632\,195\,134 \\
\hline
Total & 62\,850\,462 & 708\,622\,867 \\
\lasthline
\end{tabu}
\caption{Corpora statistics for the pre-training of the language model BERT\tsub{constr}.}
\label{tab:training_language_model}
\end{table}


We used the XLM\footnote{\url{https://github.com/facebookresearch/XLM}} tool with the Masked Language Model (MLM) objective, and with the following parameters: 6 layers, 8 attention heads and an embeddings size of 512, for a total of 34.78M parameters. In constrast, the original BERT model ``bert-base-cased'' has 12 layers, 12 attention heads and an embeddings size of 768, for a total of 110M parameters. We chose to reduce these parameters because we had less training data.

For the optimizer, we used Adam, with a learning rate equals to 0.0001, warmup steps=30K, $\beta_1$=0.9, $\beta_2$=0.999, weight decay=0.01 and $\epsilon$=000001.

We trained 
for 1016 epochs. The validation/test MLM accuracy was 53.82\%/53.97\% and the validation/test perplexity was 11.07/10.90.


\section{Experiments}

\subsection{Training process}

We trained 9 different systems by making the following parameters vary:
\begin{enumerate}[topsep=1pt,itemsep=1pt,parsep=1pt,partopsep=1pt]
    \item The training data: either \tbf{MuST-C} or \tbf{MuST-C + News Commentary}.
    \item The input language model, either \tbf{None}, \tbf{BERT\tsub{extern}} (external pretrained model ``bert-base-cased'') or \\ \tbf{BERT\tsub{constr}} (constrained on allowed data only).
    \item The embeddings size, either \tbf{512} or \tbf{1024} (only 512 when the training data is only MuST-C).
\end{enumerate}

We applied label smoothing with a parameter of $0.1$ to the cross entropy criterion (as in \citep{vaswani2017}). We trained on batches of sentences of size 100, on a single NVIDIA GTX 1080 Ti. We evaluated the quality of our system in terms of BLEU \citep{papineni2002bleu} on the development corpus at the end of every epoch, and we kept the best on a total of 250 epochs.

Finally, for the optimizer, we used Adam \citep{KingmaB14} with a fixed learning rate of $0.0001$.

\subsection{Results}

We evaluated every best system on the development corpus at the end of the training, with beam search applied with a beam size of 12.
The results on this development set and on the task's test set are in \autoref{tab:results}.

\begin{table*}[htbp]
\centering
\tabulinesep=3pt
\taburowcolors[2]3{white..gray!10}

\begin{tabu} to \linewidth {|X[3.0lm]|X[1.8cm]|X[1.5cm]|X[1.4cm]||X[1.4cm]|X[1.4cm]|X[1.4cm]|X[1.4cm]|X[1.4cm]|X[1.4cm]|}
    \firsthline
    Training data & Input LM & Embed\-dings size & BLEU (Dev) & BLEU (Test) & TER & BEER & Charac\-TER & BLEU (ci) & TER (ci) \\
    \hline
    MuST-C & None & 512 & 20.4 &  19.13 & 61.55 & 51.64 & 52.23 & 19.77 & 60.54 \\
    MuST-C & BERT\tsub{constr} & 512 & 20.5 & 20.02 & 60.64 & 51.78 & 51.26 & 20.68 & 59.53 \\
    MuST-C & BERT\tsub{extern} & 512 & 21.9 & 21.07 & 59.19 & 52.74 & 49.97 & 21.78 & 58.15 \\
    \hline
    MuST-C + News & None & 512 & 22.8 & 22.26 & 58.68 & 53.84 & 48.29 & 22.93 & 57.63 \\
    MuST-C + News & BERT\tsub{constr} & 512 & 20.4 & 20.09 & 60.90 & 51.91 & 50.66 & 20.77 & 59.79 \\
    MuST-C + News & BERT\tsub{extern} & 512 & 23.7 & 23.53 & 57.55 & 54.36 & 47.30 & 24.27 & 56.41 \\
    \hline
    MuST-C + News & None & 1024 & 23.1 & \tbf{22.72} & 58.51 & 53.94 & 48.27 & 23.47 & 57.41\\
    MuST-C + News & BERT\tsub{constr} & 1024 & 21.6 & 21.35 & 59.68 & 52.83 & 49.60 & 22.05 & 58.52 \\
    MuST-C + News & BERT\tsub{extern} & 1024 & 23.9 & \tbf{23.69} & 57.14 & 54.58 & 47.06 & 24.41 & 56.02 \\
    \lasthline
    \end{tabu}

\caption{Results of our systems on the development and the test set after 250 epochs (except in two cases, see \autoref{sec:submission}). Beam size is 12. Data are tokenized and cased.}
\label{tab:results}

\end{table*}

As we can see, in every case, using the BERT\tsub{extern} language model consistently improves the BLEU score comparing to using no language model, or using our BERT\tsub{constr} language model, by an absolute value ranging from 0.8 to 1.5 on the development set, and from 0.97 to 1.94 on the test set.

Using our BERT\tsub{constr} language model however, decreases the score comparing to using no language model, but only on the MuST-C + News dataset. When training on the \mbox{MuST-C} alone, using our language model 
adds 0.1 to the BLEU score on the development set, and 0.89 on the test set.
We think that this bad performance, compared to BERT\tsub{extern}, may be explained by one or several factors such as: not having enough training data, a suboptimal choice of hyperparameters or because we stopped the training of the LM too early. 

Concerning the training data, having more is generally better, but knowing that the MuST-C only consists of 112\,993 sentences, the final score obtained by the systems trained solely on this corpus is still considerable. We can also notice that using the MuST-C alone is where both language models are the more useful, 

Finally, using an embeddings size of 1024 instead of 512 on the second dataset is useful and it gives us our best scores, but the difference is not really high (+0.2 on the dev set and +0.16 on the test set, when using Bert\tsub{extern}).

\subsection{Submission}\label{sec:submission}

For our submission, we provided the output of our 9 systems on the test set, with the same beam size of 12, but we added an extra detokenization step at the end using the script \ttt{detokenizer.perl} provided by Moses.

Due to a lack of time, we stopped the training of some systems on less than 250 epochs. 
In the case where we use MuSTC + News as training data and with the BERT\tsub{constr} language model, we stopped the training at epoch 68 with the embeddings size of 512, and epoch 55 with the embeddings size of 1024. The BLEU score on the development corpus obtained by these models, after the training complete, are respectively 22.4 (instead of 20.4) and 22.7 (instead of 21.7).

We submitted our best constrained system as our primary submission (the one obtaining 23.1 BLEU on the dev set) and all the others as constrastive.


\section{Conclusion}

In our submission for the English-Czech Text Translation Task, we submitted a neural MT system based on the Transformer architecture, and we studied the impact of a pre-trained language model used as input embeddings.

We experimented on two training sets: one which consists of a specialized speech translation corpus only, and the other which includes also a news commentary corpus. We compared the performance of an external BERT model provided by the original authors (in unconstrained settings) and a constrained BERT model that we trained ourselves on the allowed data only.

Our results showed that our model really benefits from the external BERT model trained on more than 3 billions words, really improving the quality of the translation in every case, but our constrained BERT model trained on less than 1 billion words does not always give a useful information to the MT system.
However, we believe that this could be due to a suboptimal choice of hyperparameters (different embeddings size, optimizer, etc.) or because we stopped the training too early.


\bibliographystyle{IEEEtran}

\bibliography{biblio}

\end{document}